\documentclass[10pt,twocolumn]{article}
\usepackage[margin=2cm]{geometry}
\usepackage[utf8]{inputenc}
\usepackage[T1]{fontenc}
\usepackage{lmodern}
\usepackage{amsmath, amssymb, amsfonts}
\usepackage{graphicx}
\usepackage{microtype}
\usepackage{booktabs}
\usepackage{hyperref}
\usepackage{authblk}
\usepackage{adjustbox}
\usepackage{ulem}
\usepackage{xcolor}
\usepackage{booktabs}
\usepackage{tabularx}
\usepackage{array}
\usepackage{multirow}

\title{
  \resizebox{\textwidth}{!}{
    3L-D: A Low-Light Level Dataset from a Sensor-Agnostic, Alignment-Preserving Acquisition Framework
  }
}

\author[1]{Léo Rueff}
\affil[1]{3L-OPTRONICS, 2 rue Marc Donadille, 13013 Marseille, France}
\author[1]{Maurgène Ghezal}
\date{}
\author[1]{Valentin Gorse}
\date{}
\author[1]{Astrid Dufaure}
\date{}
\author[1]{Esteban Monsalvo}
\date{}
\author[1]{François Atteia}
\date{}
\author[1]{Ludovic Escoubas}

\date{}
\begin{document}
\twocolumn[
  \begin{@twocolumnfalse}
    \maketitle
    \begin{abstract}
       Supervised low-light image and video enhancement has achieved remarkable progress, yet current models remain fundamentally constrained by the limitations of existing datasets. Prevailing methods typically rely on uncontrolled ambient captures, camera parameter manipulation (e.g., exposure or ISO adjustments), or neutral density filters, failing to replicate authentic night-sky near-infrared spectral content and genuine low-photon noise distributions. Furthermore, capturing dynamic paired sequences without introducing warping artifacts or sensor-specific biases remains an open challenge. In this paper, we introduce \textbf{3L-Dataset}, a physically calibrated low-light image and video dataset generated using a modular, sensor-agnostic acquisition framework. Our test bench faithfully reproduces the standardized Midavaine night level scale from Level~1 (full moon, $\sim$1000~mlx) down to Level~5 (overcast moonless night, $\sim$0.1~mlx), spanning the visible to near-infrared spectrum across diverse realistic environments, including countryside, forest, urban, and desert scale models. Using high-precision programmable mechanical actuation, our acquisition framework guarantees sub-pixel spatial alignment across synchronized sextuplets, each composed of one reference image and five corresponding images acquired at the five standardized night levels, without post-processing image warping. We evaluate multiple representative deep learning baselines across varying night levels, revealing critical trade-offs between reconstruction fidelity, perceptual quality, and inference latency under severe photon starvation. The modular acquisition framework can be adapted to different cameras, optical systems, sensor technologies, and resolutions, enabling the dataset to be continuously expanded.
    \end{abstract}
    \vspace{1.5em} 
  \end{@twocolumnfalse}
]

\section{Introduction}
The recent advancements in deep learning have significantly improved the performance of low-light enhancement algorithms, yet the effectiveness of these models remains fundamentally limited by the scarcity of diverse and accurately registered training data. Capturing high-quality video in low-light conditions is essential for critical applications such as surveillance, robotics, or defense, but it is an inherently ill-posed problem due to the complex interaction between photon noise, sensor gain, and temporal artifacts. While synthetic datasets are often used to bridge the data gap, they frequently fail to replicate the authentic distortions and spatiotemporal incoherent noise found in real-world scenarios. Furthermore, existing datasets that provide real-world pairs often suffer from sensor-specific biases or registration inaccuracies, which can introduce artifacts and limit the generalizability of trained models across different imaging hardware.

Multiple low-light datasets exist, but often lack correspondence with real outdoor nighttime conditions. For instance, LOL\cite{LOL}, LSRW\cite{r2rnet} or SIDD\cite{SIDD} focus on simulating low-light conditions through camera parameter manipulation, in particular integration time and ISO. This approach leads to two key limitations: daytime images do not carry the near-infrared spectral content (780 - 1100 nm) that dominates natural nighttime illumination, and modifying camera parameters artificially alters noise distributions, inflating electronic noise independently of the actual photon flux. MIT-Adobe FiveK\cite{MIT} takes a different approach by using post-processing tools to retouch raw images, achieving a low-light appearance through tone mapping and curve adjustment rather than actual low-illumination capture, thereby retaining the noise statistics and spectral content of daylight images. SMOID~\cite{SMOID} uses neutral density (ND) filters to reduce the photon flux reaching the sensor. Although ND filters are generally designed to provide approximately uniform attenuation in the visible spectrum, their transmission may vary in the near-infrared. Consequently, they may fail to preserve the visible-to-near-infrared ratio characteristic of natural low-light conditions. The resulting illumination spectrum can therefore differ from that observed under natural nighttime environments. ND filters can also introduce color shifts and optical aberrations. More broadly, characterizing illumination using a single lux value is insufficient for high-end applications, such as aerospace or tactical vision, since lux does not describe the spectral distribution of the illumination or its visible-to-near-infrared balance. Performance should instead be validated against precise atmospheric and celestial night levels.

In this paper, we introduce a novel low-light image and video dataset developed by 3L-OPTRONICS to meet the strict demands of professional vision systems. The dataset is generated using a modular acquisition method designed to accommodate different cameras, optical systems, and sensor configurations. It is acquired under physically calibrated illumination conditions that replicate natural nighttime environments, grounded in the Midavaine night level scale \cite{midavaine}, ranging from Level~1 (full moon, clear sky, $\sim$1000~mlx) to Level~5 (moonless, overcast, $\sim$0.1~mlx), and covering the full spectral range from visible to near-infrared. This standardized photometric protocol ensures that every scene is characterized by precise and reproducible radiometric values, providing a scientifically grounded foundation for supervised learning. Fully registered sextuplet video sequences are provided across multiple sensor configurations. For each scene and temporal instant, the acquisition protocol produces a synchronized sextuplet consisting of one reference image acquired under the reference illumination (N0) and five corresponding low-light images acquired at Night levels N1 to N5.This organization is richer than conventional paired datasets, while remaining compatible with pair-based supervised learning.

The main contributions of this work are as follows:
\begin{itemize}

    \item A large-scale low-light image and video dataset acquired under physically calibrated nighttime illumination, covering the full range of Midavaine night levels and organized into aligned sextuplets.

    \item A modular acquisition framework enabling the generation of aligned sextuplet data with different cameras, optical systems, and hardware configurations, while preserving pixel-level spatial alignment.
    
    \item A diverse set of scenes matching real-world outdoor environments, including 
    countryside, forest, urban, and desert scenarios.
    
    \item Benchmark evaluations of low-light enhancement and denoising methods on the proposed dataset.
    
\end{itemize}

A representative sample of the dataset and the associated resources is publicly available at:
\url{https://3l-d.github.io/3L-Dataset/}.

\section{Dataset Overview}

\subsection{Standardized Illumination}

The low-light illumination used in this study was designed to replicate night vision conditions. This work is based on the research of T. Midavaine \cite{midavaine}, who modelled the illuminance conditions encountered on Earth, particularly at night. In this situation, the perceived illumination on the ground varies depending on various sources such as the moon, stars, the Milky Way, etc. T. Midavaine defines night levels ranging from Night 1 to Night 5, with specific sky situations for each level. The different night levels, their conditions and their associated ground illuminance values are listed in Table \ref{tab:nightlevel}.

\begin{table}[h!]
\centering
\begin{adjustbox}{width=0.48\textwidth}
\begin{tabular}{|l|c|c|}
\hline
\textbf{Night level} & \textbf{Conditions} & \textbf{Ground illuminance [mlx]} \\ \hline
1 & Full moon + clear sky & 1000 to 40 \\ \hline
2 & Full moon + cloud cover  & 40 to 10  \\ \hline
3 & Crescent moon + clear sky  & 10 to 2 \\ \hline
4 & Moonless + clear sky  & 2 to 0.7 \\ \hline
5 & Moonless + cloud cover & 0.7 to 0.1  \\ \hline
\end{tabular}
\end{adjustbox}
\caption{\label{tab:nightlevel}Night levels and conditions \cite{midavaine}}
\end{table}

\begin{figure}[h!]
\centering
\includegraphics[width=8cm]{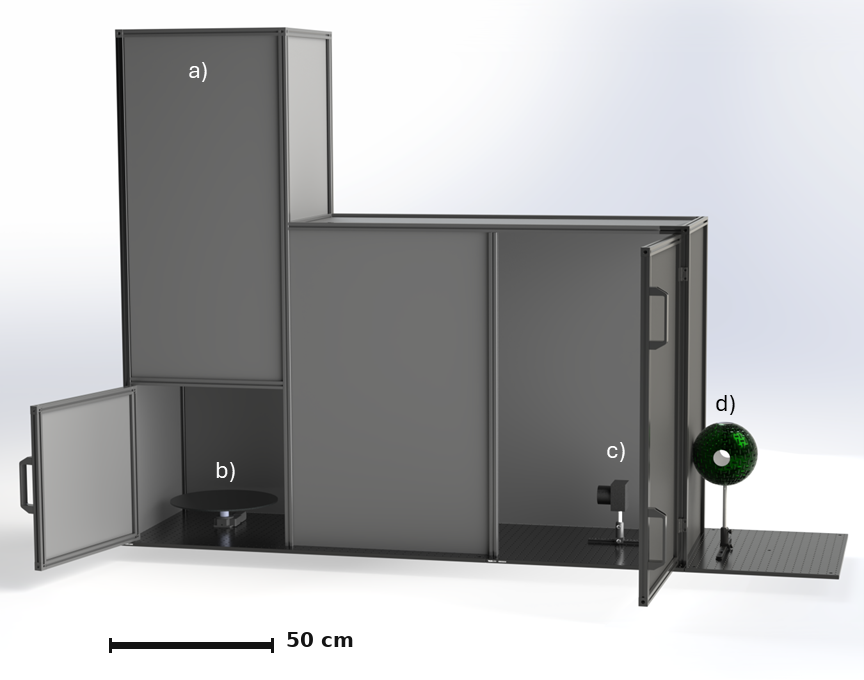}
\caption{Modelisation of our low-light dataset generation setup with : a) Light-proof enclosure, b) High-precision rotary stage, c) Acquisition system, d) External lightning system}
\label{boite_noire}
\end{figure}

Each night level is associated with a specific illuminance spectrum which we reproduce in our test bench (Fig.~\ref{boite_noire}). The illumination system used is a patented device designed to reproduce the spectrum of different night levels. The system provides uniform, diffused lighting across the scene area. A precision luxmeter and a spectrometer are used to calibrate the illuminance on the scene. The resulting image is therefore dependent on the scene's albedo. 

For each scene and temporal instant, the acquisition protocol produces a synchronized sextuplet which consists of one reference image acquired under nominal illumination and five corresponding low-light images acquired at Night levels 1 to 5. The six images share the same scene content and temporal position, providing a richer representation of the illumination transition than a conventional image pair. This organization also allows the extraction of standard reference-to-low-light pairs for supervised enhancement and restoration tasks.

\begin{figure*}[t]
    \centering
    \setlength{\tabcolsep}{2pt}

    \begin{tabular}{cccccc}
        \includegraphics[width=0.155\textwidth]{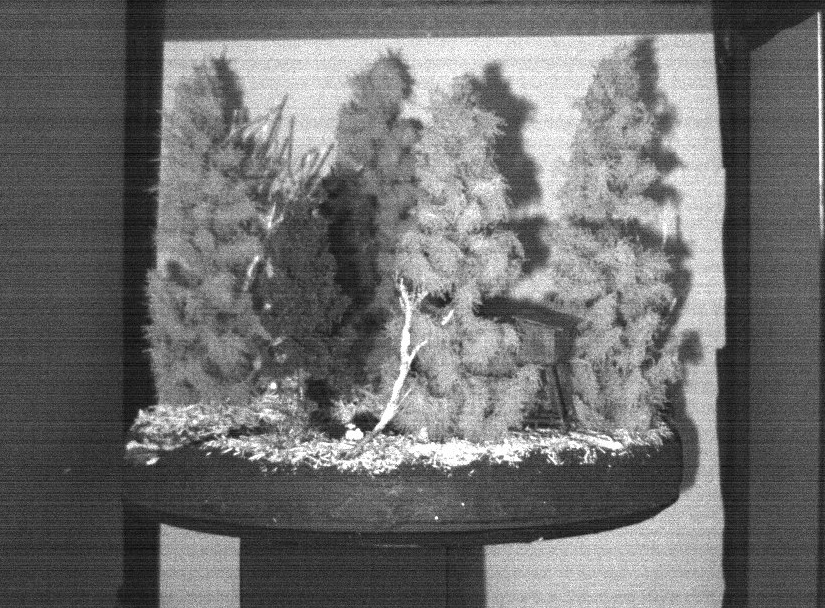} &
        \includegraphics[width=0.155\textwidth]{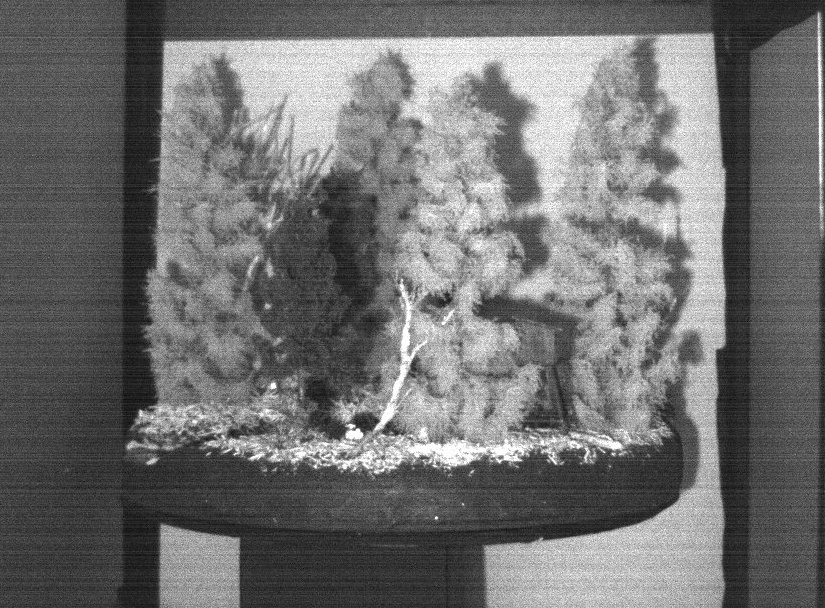} &
        \includegraphics[width=0.155\textwidth]{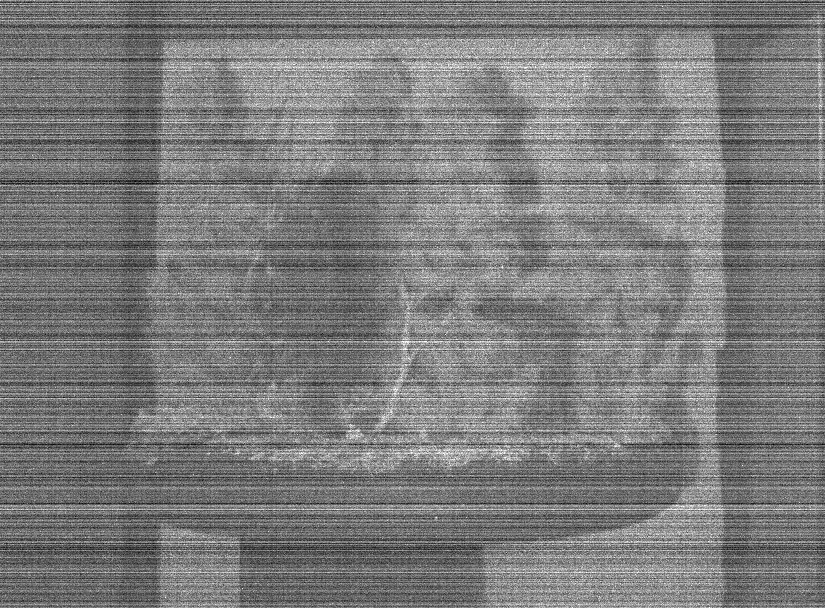} &
        \includegraphics[width=0.155\textwidth]{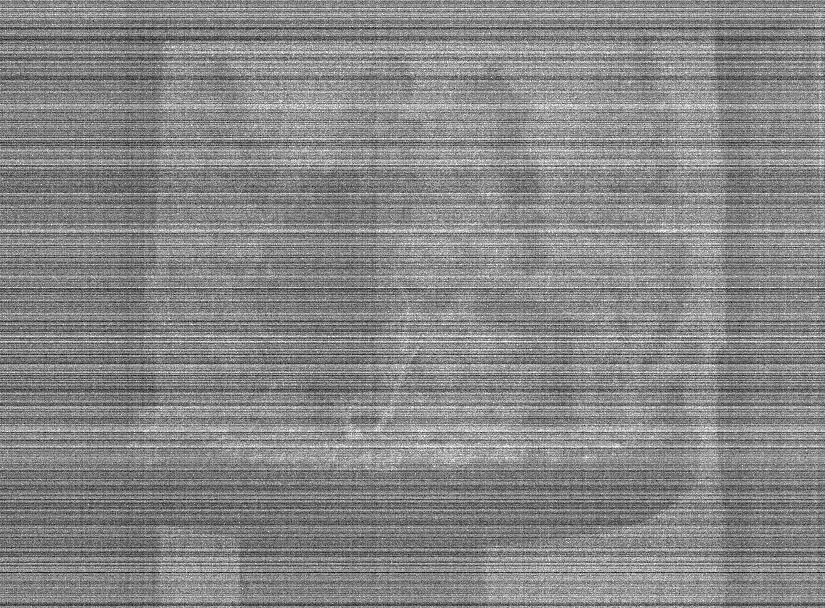} &
        \includegraphics[width=0.155\textwidth]{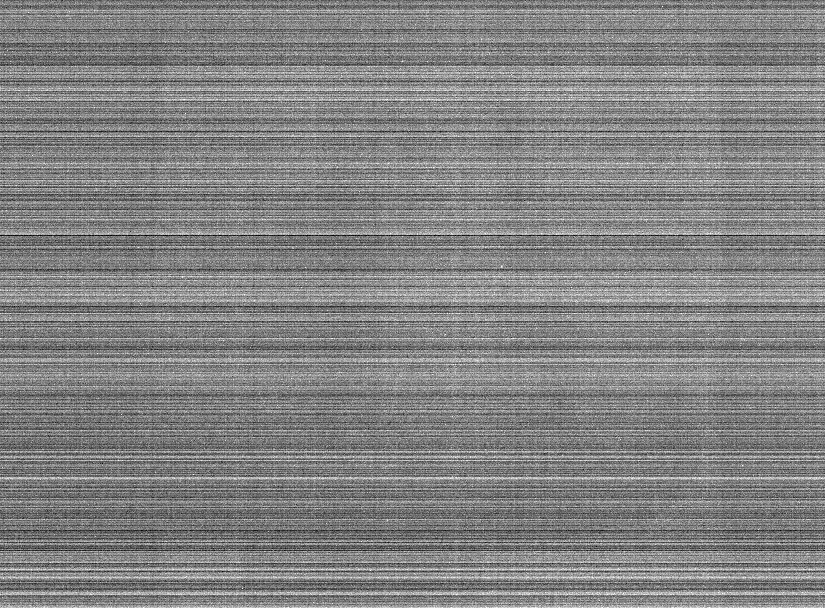} &
        \includegraphics[width=0.155\textwidth]{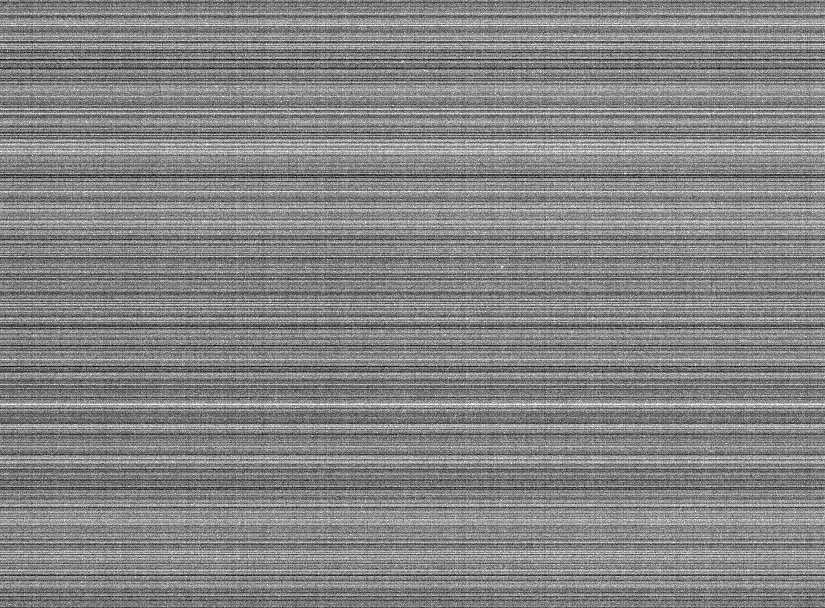} \\[-1mm]

        \textbf{\scriptsize N0} &
        \textbf{\scriptsize N1} &
        \textbf{\scriptsize N2} &
        \textbf{\scriptsize N3} &
        \textbf{\scriptsize N4} &
        \textbf{\scriptsize N5}
    \end{tabular}

    \caption{Example of an aligned sextuplet acquired from the same scene at the same temporal instant. N0 corresponds to the reference image acquired under nominal illumination, while N1 to N5 correspond to the five standardized night levels.}
    \label{fig:sextuplet_example}
\end{figure*}

To make the sextuplet structure explicit, Fig.~\ref{fig:sextuplet_example} shows six aligned images acquired from the same scene at the same temporal instant. In this representation, N0 denotes the reference image acquired under nominal illumination, while N1 to N5 correspond to the five standardized Midavaine night levels. The scene content and spatial geometry remain unchanged, while the illumination level and the associated low-light degradation progressively evolve.

\subsection{Scene Composition and Material Diversity}

\begin{figure}[h!]
\centering
\includegraphics[width=8cm]{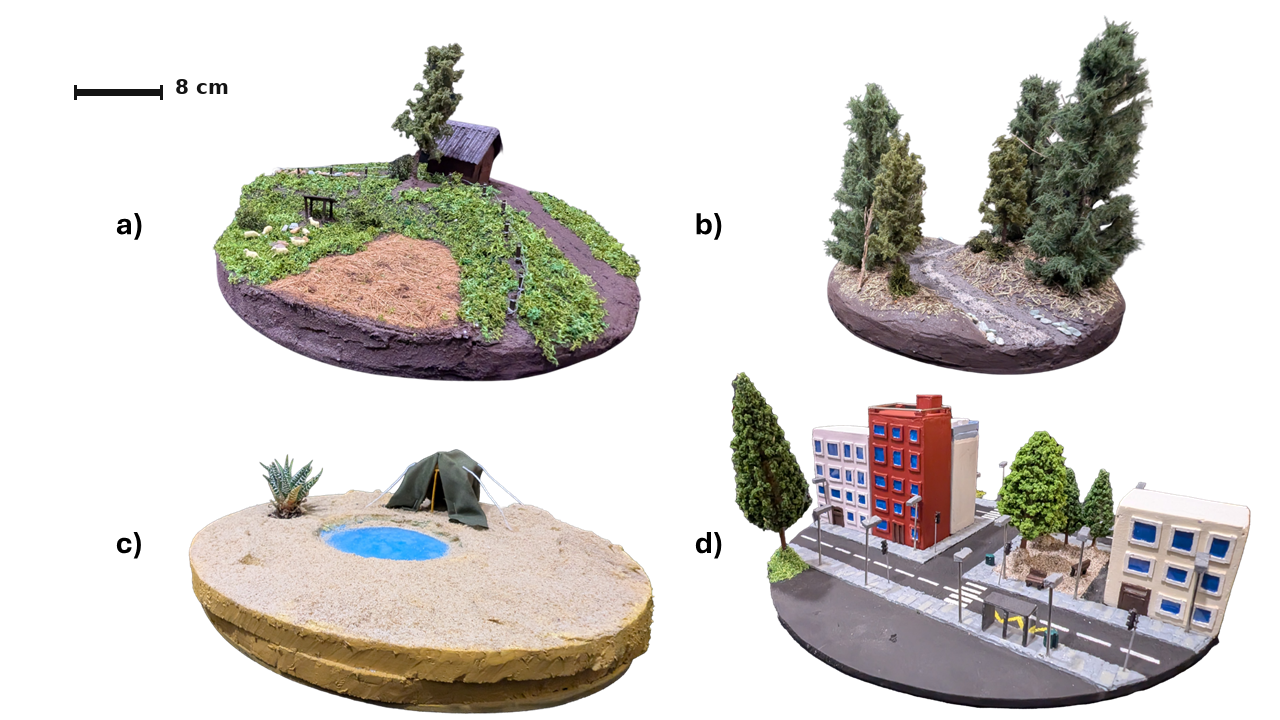}
\caption{Examples of scale models created by 3L-OPTRONICS to replicate real-world environments: a) Countryside, b) Forest, c) Desert, d) City.}
\label{figure2}
\end{figure}

In order to reproduce as faithfully as possible real-world scenarios - such as the city, the countryside, the forest, etc - scale models (Example on Fig.~\ref{figure2}) have been designed by 3L-OPTRONICS using carefully selected materials that accurately reflect the albedo as well as the textures of these environments. While the models are not composed exclusively of natural materials, the materials used have been chosen for their abiliy to closely match the albedo properties of real-world sceneries. In order to cover a wide range of situations, four models were originally built; others are currently being designed and built. 

\subsection{Camera and Sensor Diversity}

The proposed dataset generation method is designed to be compatible with a wide range of imaging systems. The modular acquisition setup allows different cameras, optical systems, sensor technologies, resolutions, and imaging configurations to be integrated without modifying the illumination protocol or the overall acquisition procedure. This flexibility makes it possible to generate comparable low-light data for various camera types, from embedded and industrial systems to high-resolution scientific or consumer cameras.

For each imaging system, the relevant camera and optical parameters can be configured while preserving the same scene content, illumination levels, and acquisition protocol. The method can therefore be adapted to different focal lengths, fields of view, sensor formats, and spectral sensitivities. This flexibility of the acquisition method facilitates the extension of the dataset to new hardware configurations and enables the study of low-light enhancement methods under a broader range of imaging conditions.

\subsection{Registration and Alignment}

A fundamental challenge in generating supervised datasets for low-light video enhancement is ensuring perfect pixel-wise alignment between the noisy low-light frames and the clear day-light ground truth. While existing methodologies often rely on post-capture image registration or non-rigid warping to correct for mechanical tolerances, such techniques can inadvertently alter low-level noise characteristics or introduce spatial artifacts. Our acquisition framework mitigates these issues by shifting the burden of registration from post-processing software to high-precision hardware complemented with software validation.

\subsubsection{Mechanical Precision and Motion Control}

To achieve repeatable motion profiles across varying illumination levels, we utilize a high-precision controlled mechanical actuation system. The platform offers an exceptional spatial resolution, allowing identical trajectories to be executed for the reference sequence and the five low-light sequences corresponding to Midavaine levels 1 to 5. This ensures that all six frames within each sextuplet remain spatially and temporally aligned. By strictly controlling the angular velocity and positioning of the scene objects, we ensure that the temporal and spatial coordinates of each frame remain consistent throughout the acquisition process, effectively decoupling the motion from the sensor's integration time.

\subsubsection{Mathematical Framework for Alignment Validation}

To rigorously quantify the mechanical registration accuracy, we employ the principle of optical flow, which operates under the brightness constancy assumption. This hypothesis suggests that the intensity of a specific pixel $I(x, y, t)$ remains constant despite a small displacement $(\Delta x, \Delta y)$ over a temporal interval $\Delta t$, leading to the fundamental constraint equation:
\begin{equation}
    \frac{\partial I}{\partial x} u + \frac{\partial I}{\partial y} v + \frac{\partial I}{\partial t} = 0
\end{equation}
where $u = dx/dt$ and $v = dy/dt$ represent the horizontal and vertical components of the flow vector, respectively. While standard differential methods are often sensitive to noise, a critical factor in low-light imaging, our framework utilizes the Farneback algorithm for dense estimation. This method approximates the local neighborhood of each pixel using a quadratic polynomial expansion:
\begin{equation}
    f(\mathbf{x}) \approx \mathbf{x}^T \mathbf{A} \mathbf{x} + \mathbf{b}^T \mathbf{x} + c
\end{equation}
By monitoring the displacement of these polynomial surfaces between the ground truth and the test sequences, we can detect even subtle mechanical drifts or micro-vibrations. To provide a comprehensive evaluation of the alignment integrity, we look beyond the standard Root Mean Square Error (RMSE). While RMSE offers a useful global average, it is intrinsically sensitive to outliers and stochastic noise spikes that frequently occur in low-SNR environments. To achieve a more representative measure of the system's mechanical performance, we prioritize the median flow magnitude and the 95th percentile (P95) of the displacement field.

The median flow magnitude serves as our primary indicator of global registration, representing the typical displacement experienced by the majority of pixels in the scene. A sub-pixel median confirms that the bulk of the image remains perfectly aligned with the ground truth. Simultaneously, we utilize the P95 as a conservative upper bound for mechanical tolerances. By ensuring that 95\% of the flow vectors fall below a specific threshold (e.g., $<0.8$ pixels), we can confidently rule out significant mechanical drifts, vibrations, or synchronization errors that might otherwise compromise the temporal consistency of the video.

Our experimental results indicate that the high-precision motion control maintains a median displacement significantly below the sub-pixel threshold. This level of registration is crucial for training state-of-the-art architectures, which are highly sensitive to spatial inconsistencies during the loss calculation phase. By providing natively registered data, our framework eliminates the need for histogram matching or warping prior to training, preserving the original sensor noise profile.

The alignment measurements reported in this section were obtained using a specific night-vision sensor configuration. The numerical values, including the median flow magnitude and the P95 displacement, are therefore dependent on the characteristics of the imaging system and may vary with the sensor pixel pitch, spatial resolution, optical configuration, and sampling geometry. Additional validation experiments are being conducted with other imagers to assess the transferability of these alignment results across different imaging systems. The alignment procedure itself remains applicable to these configurations, while the resulting displacement statistics should be interpreted with respect to the sensor and optical setup used.

\subsubsection{SNR Limitations and Structural Consistency}

While the aforementioned optical flow validation is highly effective for Midavaine levels 1 and 2, extreme low-light conditions (levels 3 to 5) present significant challenges for gradient-based registration. At these intensities, the integration of photon noise and a dominant noise floor leads to severe photon starvation, which can cause the brightness constancy constraint to fail. Consequently, direct pixel-wise validation via Farneback becomes statistically unreliable as the structural signal is masked by stochastic noise. 

To ensure dataset integrity across the entire radiometric range, our framework relies on the principle of mechanical temporal consistency. Since the high-precision platform and the optical assembly remain strictly untouched and programmatically controlled between the ground truth (100\% illumination) and the sequential captures (levels 1--5), the registration measured at higher SNR levels serves as a reliable proxy for the entire sequence. 

We assume that by establishing a sub-pixel alignment baseline at level 1 and 2, the geometric relationship between the sensor and the scene remains invariant for the subsequent levels where the signal is too low for traditional software-based verification.

\subsection{Intrinsic dimension and dataset variety}

The manifold hypothesis states that natural images, although embedded in a
very high-dimensional pixel space, concentrate near a low-dimensional manifold
whose dimension reflects the number of independent factors of variation in the
data~\cite{bengio2013representation}. This intrinsic dimension (ID) has
been shown to be a relevant descriptor of image datasets: natural images have
an ID far lower than their number of pixels, and datasets with a lower ID are
easier to learn and require fewer samples to generalise~\cite{pope2021intrinsic,konz2022intrinsic}.

The ID is also implicitly captured by generative models~\cite{stanczuk2024diffusion}
and is used to characterise the data in inverse problems~\cite{causin2025estimating}.
Moreover, image data do not necessarily lie on a single manifold: different
classes or subsets may occupy distinct manifolds of different dimensions, which
is known as the union of manifolds hypothesis~\cite{brown2022verifying}.
Estimating the ID of our dataset therefore provides three types of information:
whether the images are structured and learnable (low ID), whether some
acquisition conditions introduce additional sources of variation (increase in
ID), and whether the different subsets share the same structure or explore
different regions of the image space.

We estimated the ID for each illumination
level using two complementary estimators: TwoNN, based on the ratio of the
distances to the first two nearest neighbours~\cite{facco2017estimating}, and
the maximum-likelihood estimator (MLE) with $k=5$ neighbours~\cite{levina2004}.
As both estimators tend to underestimate the ID in high dimension and are
sensitive to noise~\cite{candelori2024robust}, our analysis relies on the
relative evolution of the ID across illumination levels rather than on its
absolute values. The analysis was carried
out on raw pixels, which characterise the data themselves, and on ResNet-18
features, which are less sensitive to global brightness and better reflect
scene content. Since all illumination levels are pixel-aligned and depict the
same scenes, any change in ID between levels can be attributed to the
acquisition conditions rather than to the content.

\begin{figure}[h!]
    \centering
    \includegraphics[width=8cm]{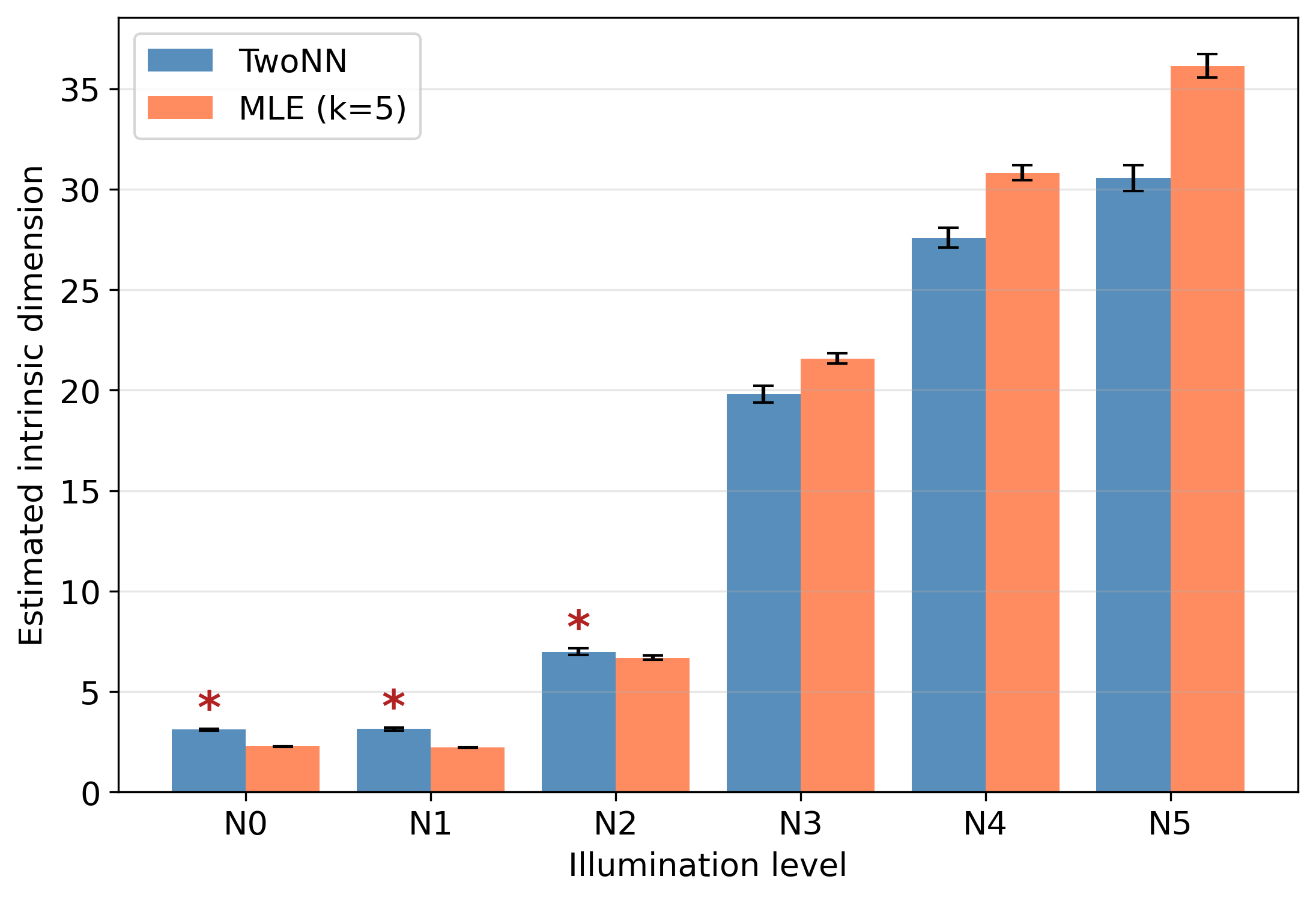}
    \caption{Intrinsic dimension per illumination level on raw pixels, estimated with TwoNN and MLE ($k=5$), from N0 (daylight reference) to N5 (darkest). Error bars give one standard deviation over 30 bootstrap samples; asterisks mark levels where the TwoNN estimate is unstable.}
    \label{fig:id_levels}
\end{figure}

\begin{figure*}[h!]
    \centering
    \includegraphics[width=\textwidth]{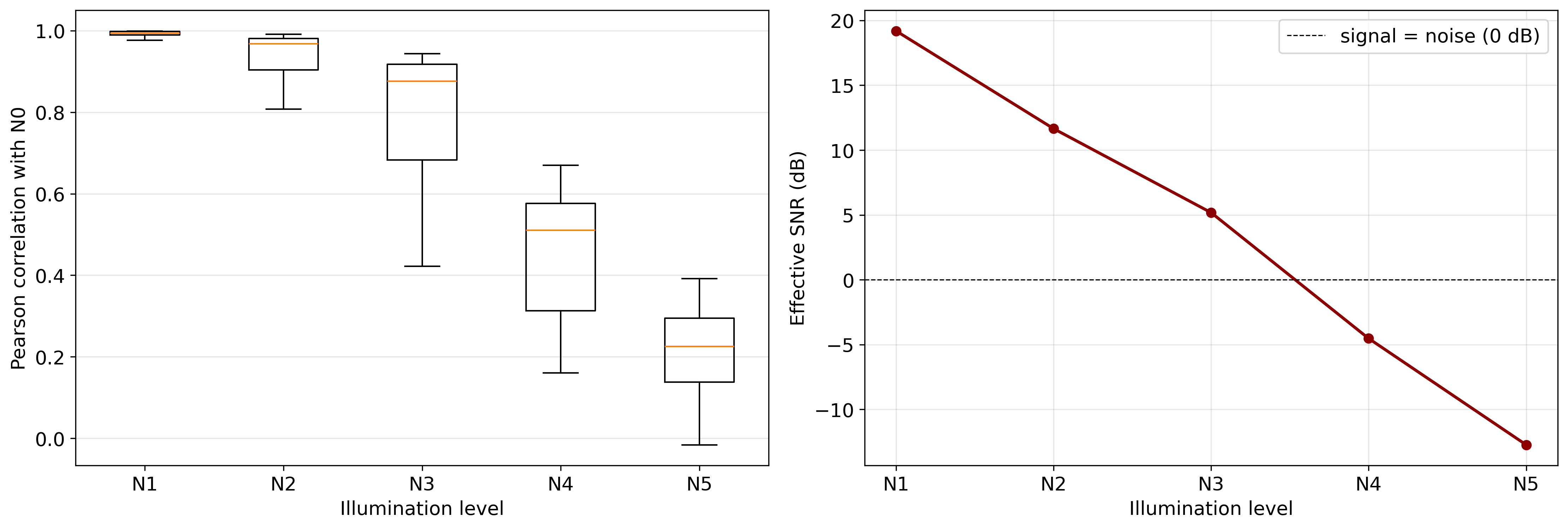}
    \caption{Content preservation per illumination level on raw pixels.
    Left: distribution of the Pearson correlation between each frame and its
    pixel-aligned daylight reference (N0). Right: effective signal-to-noise
    ratio derived from the median correlation,
    $\mathrm{SNR} = \rho^2 / (1-\rho^2)$; the dashed line marks equal signal
    and noise power (0~dB).}
    \label{fig:content_preservation}
\end{figure*}

For the reference level N0, the ID is very low (TwoNN $\approx 3.1$,
MLE $\approx 2.2$ on raw pixels) and remains unchanged at N1
(Fig.~\ref{fig:id_levels}). The clean images therefore lie on a low-dimensional manifold, consistent with the manifold hypothesis. This ID is lower than those reported for standard
image datasets~\cite{pope2021intrinsic}, which is expected for video frames whose content is strongly correlated over time. The ID then
increases sharply with decreasing illumination, reaching about 7 at N2, 20 at N3 and 31 to 36 at N5. Since all levels are pixel-aligned and depict exactly the same scenes, this increase cannot be explained by additional content: it reflects the contribution of noise, which adds random directions to the data at the finest scale, where nearest-neighbour estimators operate. This interpretation is supported by the loss of content measured in Fig.~\ref{fig:content_preservation}, which follows the same progression as the ID, and by the ResNet-18 feature space, where the same trend is observed with a smaller amplitude (MLE $\approx 25$ at N5), showing that the network
partly filters this noise.

The resulting loss of information is quantified in
Fig.~\ref{fig:content_preservation} by the correlation between each frame and
its pixel-aligned reference. The content is almost fully preserved at N1
($\rho = 0.99$, effective SNR of 19~dB) and remains dominant up to N3
($\rho = 0.88$, 5~dB). The SNR crosses 0~dB between N3 and N4: at N4
($\rho = 0.51$, $-4.5$~dB) and N5 ($\rho = 0.23$, $-12.7$~dB), noise prevails
and most of the scene content is lost, while the wider spread of the
correlations shows that frames are not degraded uniformly. This progressive
loss of content mirrors the increase in ID observed in
Fig.~\ref{fig:id_levels}.

The diversity of the content was also studied on the reference level N0
(analysis not detailed here). Each sequence taken individually has
a low ID, as expected for video data in which consecutive frames are strongly
related, whereas frames from different sequences are typically 2.44 times further apart (ratio of median distances) than frames from the same sequence (ResNet-18 features). The
different sequences therefore explore distinct regions of the image space,
indicating that the dataset covers varied scene content.

Overall, this analysis shows that the clean images of our dataset lie on a
low-dimensional manifold, which supports the manifold hypothesis and indicates
a structured, learnable distribution. Decreasing illumination does not add
diversity to the data: it degrades the content through noise, which
artificially increases the ID, and from N4 onward most of the scene content is
lost. No evidence of a union of distinct manifolds was found, since the
differences in ID between levels are explained by noise rather than by
distinct underlying structures~\cite{brown2022verifying}.

\section{Benchmarks}

\subsection{Comparison With Existing Datasets}

Table~\ref{tab:extended_lowlight_datasets} compares the proposed dataset against 
existing low-light benchmarks. Most datasets rely on uncontrolled real-world 
illumination~\cite{BVI, DUS, DRV, RViDeNet, SID, LOL, VE-LOL, DarkVision, ELD, 
LDV, Ev-LowLight}, without specifying the photometric conditions of the capture, while others use neutral density filters~\cite{SMOID, SDSD, DID}, which introduce distortions and do not match any physically relevant night level. A common limitation of dynamic datasets is the absence of alignment: DUS\cite{DUS} and DRV\cite{DRV} provide no ground truth for their dynamic sequences, and ND filter-based datasets achieve alignment only by restricting motion to a single degree of freedom. The proposed dataset addresses these shortcomings simultaneously: it is the only benchmark based on the Midavaine night level scale, providing standardized and reproducible radiometric conditions covering the full range of natural nighttime illumination. It further provides aligned sextuplet sequences with pixel-level registration under dynamic conditions. The proposed acquisition method can be adapted to multiple camera, lens, and sensor configurations, allowing new hardware and scene compositions to be integrated on demand. This makes the dataset a continuously expandable resource for low-light research.

\begin{table*}[t]
    \centering
    \caption{Comprehensive comparison of video and image datasets in low-light 
    conditions. S and D denote static and dynamic scenes, respectively. 1D 
    indicates a single type of motion. D* indicates \textit{stop motion}.}
    \label{tab:extended_lowlight_datasets}
    \resizebox{\textwidth}{!}{
    \begin{tabular}{llccccc}
        \toprule
        \textbf{Name} & \textbf{Light} & \textbf{Motion} & \textbf{Images/Videos} & \textbf{Resolution} & \textbf{Ground Truth} & \textbf{Alignment} \\
        \midrule
        BVI-LOWLIGHT\cite{BVI}     & Real      & S     & 13,200  & 4240$\times$2832  & $\checkmark$ & $\checkmark$ \\
        DUS (Static)\cite{DUS}     & Real      & S     & 3,630   & 2160$\times$1280  & $\checkmark$ & $\checkmark$ \\
        DUS (Dynamic)\cite{DUS}    & Real      & D     & 15,000  & 2160$\times$1280  & $\times$     & $\times$     \\
        DRV (Static)\cite{DRV}     & Real      & S     & 20,847  & 5496$\times$3672  & $\checkmark$ & $\checkmark$ \\
        DRV (Dynamic)\cite{DRV}    & Real      & D     & 2,397   & 5496$\times$3672  & $\times$     & $\times$     \\
        SMOID\cite{SMOID}          & ND Filter & 1D    & 35,800  & 1800$\times$1000  & $\checkmark$ & $\checkmark$ \\
        RViDeNet\cite{RViDeNet}    & Real      & D*    & 385     & 1920$\times$1080  & $\checkmark$ & $\checkmark$ \\
        SDSD\cite{SDSD}            & ND Filter & 1D    & 37,500  & 1920$\times$1080  & $\checkmark$ & $\checkmark$ \\
        DID\cite{DID}              & ND Filter & 1D    & 41,038  & 2560$\times$1440  & $\checkmark$ & $\checkmark$ \\
        SID\cite{SID}              & Real      & S     & 5,094   & 4240$\times$2832, 6000$\times$4000 & $\checkmark$ & $\checkmark$ \\
        LOL\cite{LOL}              & Real      & S     & 500     & 600$\times$400    & $\checkmark$ & $\checkmark$ \\
        VE-LOL\cite{VE-LOL}        & Real      & Mixed & >2,500  & 1080$\times$920   & $\checkmark$ & Mixed        \\
        DarkVision\cite{DarkVision}& Real      & D/S   & 320,667 & 2464$\times$2056, 2448$\times$2048, 2048$\times$2048 & $\checkmark$ & $\checkmark$ \\
        ELD\cite{ELD}              & Real      & S     & 240     & Various           & $\checkmark$ & $\checkmark$ \\
        LDV\cite{LDV}              & Real      & D     & 115     & 1920$\times$1080  & $\checkmark$ & Mixed        \\
        Ev-LowLight\cite{Ev-LowLight}& Real    & D     & >15,000 & 346$\times$260    & $\checkmark$ & $\checkmark$ \\
        \midrule
        Ours                       & Standardized night level & D    & >5,000  & Various & $\checkmark$ & $\checkmark$ \\
        \bottomrule
    \end{tabular}
    }
\end{table*}

\subsection{Representative LLIE Models Assessment}

To evaluate the effectiveness of our dataset for low-light image enhancement, we conducted a benchmark involving several representative deep learning architectures. The selected methods cover a range of enhancement paradigms, including lightweight enhancement networks, GAN-based approaches, Retinex-based models, and recent state-space or transformer-based restoration networks. Specifically, we evaluated SCI \cite{SCI}, EnlightenGAN \cite{enlightengan}, Zero-DCE \cite{zero-dce}, KinD \cite{kind}, URetinex-Net++ \cite{uretinex}, WaveMamba \cite{wavemamba}, MambaIR \cite{mambair}, and RetinexFormer \cite{retinexformer}.

\medskip
\noindent\textbf{Training setup.}
\par\smallskip

Although the dataset is organized into aligned sextuplets, supervised training was performed using conventional image pairs extracted from each sextuplet. For each night level, a pair was formed between the reference image and its corresponding low-light image. This protocol preserves compatibility with the original training procedures of the evaluated methods while taking advantage of the richer multi-level structure of the dataset.
All evaluated networks were trained using the proposed dataset on a single NVIDIA GeForce RTX 4090 GPU. The publicly available implementations were adapted only when necessary to support the organization and input format of our dataset. The original architectures, method-specific loss functions, and enhancement principles were preserved to remain faithful to the original methods. Inference times were measured on the same GPU for all models.
For the benchmark presented in this paper, the data were acquired using a high-performance night-vision sensor. The quantitative evaluation therefore focuses on Night levels N3 to N5, which provide the most challenging conditions for this imaging system while remaining representative of severe low-light enhancement scenarios.
\medskip

\begin{table*}[t]
    \centering
    \small

    \caption{Evaluation of representative low-light image enhancement networks trained on the proposed dataset at different night levels. The purpose of this benchmark is to assess the applicability of the dataset and the consistency of the results across illumination levels, rather than to rank the evaluated methods. Higher values are better for PSNR and SSIM, whereas lower values are better for LPIPS, NIQE, and inference time.}

    \label{tab:benchmark_results}

    \setlength{\tabcolsep}{3.5pt}
    \renewcommand{\arraystretch}{1.2}

    \begin{tabularx}{\textwidth}{
    >{\raggedright\arraybackslash}p{3.0cm}
    c
    c
    c
    c
    c
    c
    >{\centering\arraybackslash}X
}

        \toprule

        \textbf{Method} &
        \textbf{Year} &
        \textbf{Night level} &
        \textbf{PSNR (dB) $\uparrow$} &
        \textbf{SSIM $\uparrow$} &
        \textbf{LPIPS $\downarrow$} &
        \textbf{NIQE $\downarrow$} &
        \textbf{Inference time (ms) $\downarrow$} \\

        \midrule

        \multirow{3}{*}{\shortstack[l]{KinD\\\cite{kind}}}
        & \multirow{3}{*}{2019}
        & N3 & 41.98 & 0.9298 & 0.0277 & 16.01 & 87.22 \\

        & &
        N4 & 39.87 & 0.9232 & 0.0584 & 18.47 & 87.22 \\

        & &
        N5 & 38.41 & 0.9051 & 0.0766 & 19.96 & 87.22 \\

        \midrule

        \multirow{3}{*}{\shortstack[l]{Zero-DCE\\\cite{zero-dce}}}
        & \multirow{3}{*}{2020}
        & N3 & 37.90 & 0.8693 & 0.7785 & 10.47 & 26.21 \\

        & &
        N4 & 37.71 & 0.8678 & 0.7769 & 10.57 & 26.17 \\

        & &
        N5 & 37.67 & 0.8671 & 0.7738 & 10.71 & 26.19 \\

        \midrule

        \multirow{3}{*}{\shortstack[l]{EnlightenGAN\\\cite{enlightengan}}}
        & \multirow{3}{*}{2021}
        & N3 & 36.66 & 0.8676 & 0.0801 & 24.04 & 58.14 \\

        & &
        N4 & 36.55 & 0.8648 & 0.0803 & 26.82 & 58.14 \\

        & &
        N5 & 36.53 & 0.8642 & 0.0804 & 26.73 & 58.14 \\

        \midrule

        \multirow{3}{*}{\shortstack[l]{SCI\\\cite{SCI}}}
        & \multirow{3}{*}{2022}
        & N3 & 33.21 & 0.7698 & 0.1571 & 11.59 & 2.12 \\

        & &
        N4 & 33.08 & 0.7656 & 0.1579 & 11.63 & 1.49 \\

        & &
        N5 & 32.73 & 0.7536 & 0.1610 & 12.78 & 1.49 \\

        \midrule

        \multirow{3}{*}{\shortstack[l]{RetinexFormer\\\cite{retinexformer}}}
        & \multirow{3}{*}{2023}
        & N3 & 44.64 & 0.9687 & 0.0210 & 17.78 & 541.14 \\

        & &
        N4 & 42.18 & 0.9595 & 0.0277 & 19.70 & 541.45 \\

        & &
        N5 & 37.73 & 0.9001 & 0.0322 & 21.20 & 541.49 \\

        \midrule

        \multirow{3}{*}{\shortstack[l]{MambaIR\\\cite{mambair}}}
        & \multirow{3}{*}{2024}
        & N3 & 51.85 & 0.9614 & 0.0011 & 21.36 & 9199.43 \\

        & &
        N4 & 50.79 & 0.9572 & 0.0012 & 22.36 & 9202.46 \\

        & &
        N5 & 50.02 & 0.9505 & 0.0014 & 21.75 & 9202.87 \\

        \midrule

        \multirow{3}{*}{\shortstack[l]{WaveMamba\\\cite{wavemamba}}}
        & \multirow{3}{*}{2024}
        & N3 & 54.80 & 0.9808 & 0.0013 & 21.71 & 1322.63 \\

        & &
        N4 & 52.32 & 0.9725 & 0.0015 & 22.43 & 1321.06 \\

        & &
        N5 & 50.80 & 0.9639 & 0.0016 & 22.02 & 1318.58 \\

        \midrule

        \multirow{3}{*}{\shortstack[l]{URetinex-Net++\\\cite{uretinex}}}
        & \multirow{3}{*}{2025}
        & N3 & 40.95 & 0.9158 & 0.0311 & 20.87 & 254.01 \\

        & &
        N4 & 38.41 & 0.8976 & 0.0342 & 22.04 & 253.37 \\

        & &
        N5 & 37.44 & 0.8813 & 0.0349 & 22.44 & 253.46 \\

        \bottomrule

    \end{tabularx}
\end{table*}

Table~\ref{tab:benchmark_results} reports the results obtained by training several representative state-of-the-art low-light enhancement architectures on the proposed dataset. The purpose of this benchmark is not to establish a ranking between the evaluated methods, but to verify that the dataset can be effectively used with different enhancement paradigms. All the networks were successfully trained using the proposed data, and the obtained results remain coherent across the different night levels.

As the illumination decreases from N3 to N5, PSNR and SSIM generally decrease, reflecting the increasing difficulty of the enhancement task under more severe low-light conditions. At the same time, the models maintain meaningful performance without collapsing, indicating that the dataset provides a well-calibrated level of difficulty. The proposed data are therefore sufficiently informative to enable effective supervised training, while remaining challenging enough to reveal the progressive degradation associated with lower illumination levels. Inference times are reported for completeness and are not intended as a comparative ranking of the evaluated architectures.

\section{Conclusion}

In this paper, we introduced a novel low-light image and video dataset designed to address the fundamental limitations of existing benchmarks. Unlike prior works that simulate nighttime conditions through camera parameter manipulation, post-processing, or neutral density filters, the proposed dataset is grounded in physically calibrated illumination conditions based on the Midavaine night level scale, covering the full 
range of natural outdoor nighttime environments with a spectral range from visible to near-infrared.

The modular acquisition method can accommodate a wide range of cameras, optical systems, and hardware configurations, while providing registered sextuplet sequences under dynamic scene conditions. At each temporal instant, a sextuplet contains one reference frame and five corresponding low-light frames acquired at standardized night levels. This richer organization preserves compatibility with conventional paired training while enabling multi-level analysis of low-light degradation. This richer organization preserves the advantages of conventional paired training while enabling multi-level analysis of low-light degradation. Scene diversity is ensured by scale models replicating real-world outdoor environments, including countryside, forest, urban, and desert scenarios. The registration and alignment pipeline guarantees pixel-level spatial consistency across all acquisitions, validated through a rigorous mathematical framework and subject to explicit SNR limitations. Benchmark evaluations of state-of-the-art low-light enhancement and denoising methods confirm the challenging nature of our dataset across progressive degradation of low light levels (N3 to N5). Crucially, the noticeable performance drop observed across all evaluated baselines as night levels deepen underscores the benchmark's ability to expose current architectural trade-offs between visual fidelity, perceptual consistency, and computational latency.
The benchmark also indicates that lightweight or weakly supervised approaches, such as SCI, EnlightenGAN, and Zero-DCE, may require further adaptation when applied to physically captured scenes with complex content and realistic photon-starvation noise. Their limited performance on the proposed dataset suggests that results obtained on simpler low-light benchmarks, such as LOL, may not directly transfer to more challenging acquisition conditions. This observation further highlights the importance of evaluating enhancement methods on datasets that reproduce realistic illumination, sensor noise, and spectral characteristics.

The modular design of the acquisition setup allows the dataset to be continuously expanded with new sensors, configurations and scene compositions. 

\newpage

\bibliography{bib}
\end{document}